\documentclass[twoside,11pt]{article}
\usepackage{fullpage}

\usepackage{hyperref}       

\usepackage{authblk}
\usepackage{lmodern}
\usepackage{theapa}
\usepackage{rotating}
\usepackage{pdflscape} 
\usepackage{booktabs} 
\usepackage[utf8]{inputenc} 
\usepackage[T1]{fontenc}    
\usepackage{url}            
\usepackage{amsfonts}       
\usepackage{nicefrac}       
\usepackage{microtype}      
\usepackage{multirow,array}
\usepackage{subfigure}
\usepackage{csvsimple}
\usepackage{xcolor}
\usepackage{gensymb} 
\usepackage{graphicx}
\usepackage{amsmath}
\usepackage{dsfont}
\usepackage{multirow} 
\usepackage{amssymb} 

\newcommand{\alg}{\textsc{ALG}}
\newcommand{\mX}{\mathcal{X}}

\newcommand{\mI}{\mathcal{I}}

\newcommand{\mY}{\mathcal{Y}}

\newcommand{\mD}{\mathcal{D}}
\newcommand{\mR}{\mathcal{R}}

\newcommand{\mS}{\mathcal{S}}

\newcommand{\ind}{\mathds{1}}

\newcommand{\R}{\mathbb{R}}

\citepunct{}{\&}{and}{, }{; }{, }{}{}{.}
\newcommand{\citep}[1]{(\cite{#1})}
\newcommand{\citet}[1]{\citeauthor{#1}~(\citeyear{#1})}

\numberwithin{claim}{section} 
\numberwithin{observation}{section} 
\numberwithin{question}{section}

\newcommand\abs[1]{\left| #1  \right|}

\usepackage[export]{adjustbox}

\begin{document}
\title{Predicting Strategic Behavior from Free Text}

\author[1] {Omer Ben{-}Porat}
\author[1] {Sharon Hirsch}
\author[1] {Lital Kuchy}
\author[2]{Guy Elad}
\author[1]{Roi Reichart}
\author[1]{Moshe Tennenholtz}
\affil[1]{Industrial Engineering and Management, Technion---Israel Institute of Technology}
\affil[2]{Computer Science, Technion---Israel Institute of Technology}
\affil[ ]{}
\affil[ ]{\texttt{\{omerbp,sharonhi,slitalku,sguyelad\}@campus.technion.ac.il}\\ \texttt{\{roiri,moshet\}@technion.ac.il }}

\maketitle
\begin{abstract}
The connection between messaging and action is fundamental both to web applications, such as web search and sentiment analysis, and to economics. However, while prominent online applications exploit messaging in natural (human) language in order to predict non-strategic action selection, the economics literature focuses on the connection between structured stylized messaging to strategic decisions in games and multi-agent encounters. This paper aims to connect these two strands of research, which we consider highly timely and important due to the vast online textual communication on the web. Particularly, we introduce the following question: Can free text expressed in natural language serve for the prediction of action selection in an economic context, modeled as a game? 

In order to initiate the research on this question, we introduce the study of an individual's action prediction in a one-shot game based on free text he/she provides, while being unaware of the game to be played. We approach the problem by attributing commonsensical personality attributes via crowd-sourcing to free texts written by individuals, and employing transductive learning to predict actions taken by these individuals in one-shot games based on these attributes. Our approach allows us to train a single classifier that can make predictions with respect to actions taken in multiple games. In experiments with three well-studied games, our algorithm compares favorably with strong alternative approaches. In ablation analysis, we demonstrate the importance of our modeling choices---the representation of the text with the commonsensical personality attributes and our classifier---to the predictive power of our model.
\end{abstract}

\maketitle
\section{Introduction} 
\label{sec:intro}

Much online activity is centered around text written by people. People send messages through a wide variety of communication media including email, Whatsapp, SMS, blogs and Facebook pages. In many cases these texts are connected to future actions, e.g., a request for some course of action or activity scheduling, among other alternatives. Moreover, web-search is probably the most successful tool which has emerged in the Web. In this context, one aims to predict an action of an individual, i.e., whether he will click on a particular page, from a text this individual provides. Web-search is far from being the only example of a powerful text-based action prediction tool. Another such prominent application is sentiment analysis \citep{pang2002thumbs,pang2008opinion}, that provides a valuable signal for the prediction of an individual's choice among alternatives (see, e.g., \cite{ghose2010estimating,ravi2015survey}). This signal is particularly useful when extracting opinions from textual reviews, that are abundant across the Web and have a strong impact on purchase decisions of wide crowds.

The connection between messaging and intention signaling to action is central also in economics. Indeed, the 2001 Nobel prize in economics was presented to Akerlof, Spence, and Stiglitz, for their pioneering lines of research, showing how signaling of information can alter strategic interactions (see \cite{Spence}).
Of particular interest is the study of cheap talk \citep{CrawfordSobel}, where messaging may include arbitrary communication about private information which is carried out prior to the action that determines the parties' payoff, that is, the economic outcome is independent of the messages transmitted prior to action selection. An interesting variant of this setting is pre-play communication \citep{Rabin}, where economic agents communicate before playing in a non-cooperative game. Typically, such communication refers to private information where the agents might lie, or to a declaration of an intention to choose an action. Importantly, it is only the game itself and not the pre-game messaging that affect the outcome. Notice that studies of cheap talk and pre-play communication typically use a structured language: Participants communicate through announcements on planned behavior or available information, which are elements of the game, rather than through arbitrary free text.

Given the above, it is evident that the connection between messaging and action is fundamental to both central applications in the web as well as to economics. The major differences are obvious as well: While the major online applications focus on the study of the connection between messaging in natural language to (non-strategic) action selection, the economics literature focuses on the connection between structured stylized messaging to strategic decisions in games and multi-agent encounters. It is therefore natural to ask: Can free text expressed in natural language serve for the prediction of an action to be selected in an economic context, modeled as a game? Indeed, the main aim of this paper is to initiate such a study. Moreover, text in online media need not necessarily be expressed in the context of a given game or action selection. Therefore, it may be even more intriguing to understand whether free text provided by individuals who are unaware of the games to be played, may induce a useful signal for the prediction of action selection in a game.

In service of the above, we introduce an experimental setup consisting of two steps. In the first step, individuals were asked to provide free text with some personal story. They were told there would be a second step, but did not know what would be its structure. In the second step, they were matched to play three one-shot games (with no communication). These games were classical one-shot games, namely a pure coordination game, a congestion game, and a classical non-cooperative game (\textit{Door, Box} and \textit{Chicken}, correspondingly, see Figure \ref{fig:normal form games}). The aim was to study whether one can predict the individuals' actions in a particular game based on their provided texts. 

The approach we have taken to tackle the above challenge is as follows. We created a description of the texts through commonsensical personality attributes.  In order to make a sound treatment, we asked a group of students to reach a consensus on a set of attributes, and used crowd-sourcing in order to annotate the texts according to these attributes. Given these, we employed an ML technique known as transductive learning \citep{gammerman1998learning,joachims1999transductive,joachims2003transductive}. The characteristic property of transductive learning is that at training time, the algorithm is exposed to all examples (i.e., all written texts), but to the labels (i.e., actions selected) of only a subset of these examples. The goal of the algorithm is then to infer the unknown labels from the known ones. An elegant property of our particular method is that we point to one structure that has predictive power with respect to different games. This is achieved by clustering over the input features (text-based personality attributes) while ignoring the labels. Notice that as choice prediction based on text is typically employed in settings where we already have the population texts (e.g., people share info in blogs, email, etc.) before we need to predict their actions, transductive learning is a natural setup. It is not that a new individual is born when we are called to make a prediction---we can use the texts he/she wrote in the past. 

Our aim was to test whether we can outperform in our predictions a standard majority benchmark, where the action predicted by an individual in the test set is taken as the most popular action of individuals in the entire data set. Our results are encouraging and show that indeed an individual action prediction in strategic situations may be performed based on free text he/she provided. Moreover, in ablation analysis we demonstrate the contribution of our specific modeling choices: employing a clustering algorithm and representing the text with personality attributes of their authors. We further demonstrate the increased predictive power of our proposed set of personality attributes compared to a previously proposed set that includes the famous big five personality dimensions \citep{john1991big}.
Needless to say that this study is mainly a call for action. Large scale experiments will be needed to test the significance of our findings.

The rest of the paper is organized as follows. Section \ref{sec:previous} discusses previous work. While, to the best of our knowledge, we are the first to address the task at hand, we survey previous work on machine learning for action prediction in strategic-form games and on natural language processing for text-based prediction.  Section \ref{sec:task} defines our task, including the games we consider,  text-based prediction with personality attribute representation, and the transductive learning setup. Section \ref{sec:data} describes our data including the data collection procedure, and provides qualitative analysis of the collected texts as well as game performance statistics. Section \ref{sec:approach} describes our modeling approach, including the algorithmic details of the transductive clustering-based classifier, as well as the novel set of personality attributes and its collection through crowd-sourcing. Section \ref{sec:experiments} provides the details of our experiments, including the baseline models to which we compare our approach and the evaluation measures. Finally, Sections \ref{sec:results} and \ref{sec:discussion} discuss our results and the derived conclusions.

\section{Related Work}
\label{sec:previous}

Machine learning (ML) has been applied in the past to predict human actions, both based on previous actions and from text. This previous work, however, is substantially different from ours. In this section, we discuss those differences and crystallize our novel contributions.

\subsection{Machine Learning for Action Prediction in Strategic-Form Games}

While our setup refers to action prediction in a single one-shot strategic-form game, previous work successfully employed ML techniques in service of action prediction in an ensemble of games. Indeed, the first work to use machine learning techniques for action prediction in one-shot strategic-form games we are aware of is \citet{ABT}. This work deals with an ensemble of games rather than a single game, with the aim of learning an individual's choice in a given game based on population behavior in the whole ensemble of games, and the individual's choice in the other games. Interestingly, the benchmark indicated above, where the predicted action is the most popular action in the game of interest, does defeat in that context leading experimental economics procedures based on cognitive models \citep{Camerer,CG}, as well as machine learning techniques relying on large data sets, but is outperformed by a simple learning method based on association rules. 

The fact that modern ML techniques based on deep neural networks outperform models based on cognitive hierarchies when the former are trained on an ensemble of games, has been illustrated recently in \citet{HartfordWL16}. 
In \citet{PlonskyEHT17}, the authors show for the first time how psychological features can be integrated with Machine Learning techniques in service of choice prediction. The games in their work are games against nature (i.e., choice among gambles) as is common in the psychology literature. 

Notice that all the above techniques heavily base their choice prediction on having access to the way the agents chose their actions in other games. Particularly, they fall into one of two settings. Works that address the first setting (see, e.g., \cite{ABT}, and the references therein), try to predict the behavior of individuals. They encode every individual by her play in several labeled games (i.e., all the individuals have played the same games in the past), and the predicted variable is the behavior of that individual in a new, unseen game. Works that address the second setting (e.g., \cite{HartfordWL16,PlonskyEHT17}), do not care about individual predictions. Every ``point'' is a choice problem, e.g., a selection between two lotteries that are encoded by probabilities and rewards, and its label is the population statistics. In particular, these works aim to predict the statistics (e.g., mean, variance, etc.) of the predictions people make on this data point, and this is different from predicting the behavior of one individual. Consequently, given a new choice problem, with unseen probabilities and rewards, the prediction is aimed at the population and is the same for all individuals.

In contrast, in the setting we address in this paper, we aim to predict the behavior of each individual in a new game, but we do not learn from previous plays. In order to address this challenging task, we exploit texts written by individuals and attempt to map texts to actions in a given game.
\subsection{Natural Language Processing for Text-based Prediction}

Researchers in Natural Language Processing (NLP) and ML have extensively explored text-related prediction tasks. The most basic tasks have to do with predictions of basic properties of the text itself such as its author, topic, and sentiment \citep{joachims1999transductive,pang2002thumbs,steyvers2007probabilistic,pang2008opinion,gill2009they}. Text-based prediction, however, goes beyond such tasks. For example, researchers tried to figure what people will think about a given text, e.g., predicting the citation count of a scientific paper \citep{yogatama2011predicting,sim2015utility}. A more ambitious effort is drawing predictions about real-world events based on texts that discuss related information \citep {smith2010text}. Examples of such efforts include the prediction of risk from financial reports \citep{kogan2009predicting}, the prediction of the revenues a movie is about to yield from the text of its reviews \citep{joshi2010movie} and predicting the polls (election outcomes) from related tweets \citep{o2010tweets}.

Two strands of the literature on text-based prediction are most related to our efforts. Firstly, several works aimed to draw predictions about future actions of the authors of given texts, in cases where the texts are directly related to the actions. For example, \citet{niculae2015linguistic} tried to predict actions in an online strategy game based on the language produced by the players as part of the inter-player communication required in the game. Secondly, some works tried to infer properties of an author's character \citep{bamman2013learning,gill2009they,golbeck2011predicting} or of his/her emotional state \citep{eichstaedt2018facebook} from sources of text such as social media posts, blogs and tweets, as well as from descriptive text, such as movie plot summaries. This line of research builds on empirical and theoretical research in psychology that connects the language people use and their personality (e.g., \cite{fast2008personality,hirsh2009personality,yarkoni2010personality}). These works, however, do not aim to predict human actions based on their personality attributes as reflected in the text, and even if such a prediction is made, it is tightly connected to the text or the social media platform on which the text has been posted \citep{mahmud2013recommending,lee2014will}.

In this paper we take a step forward, aiming to draw text-based predictions about actions that are not discussed or even implied in the text---the author is even unaware of the game he/she is going to play after writing the text. Instead, we ask our participants to write a personal text about a previous meaningful experience they are willing to share, and try to predict their actions in unrelated strategic situations. In order to make the desired action predictions, we extract attributes of the writer's character, that we believe can help in the main action prediction task. Our approach is very much related to the framework of learning with latent variables \citep{koller2009probabilistic}, where the writer's personality attributes are auxiliary variables that should help the main action prediction task. Note however, that as opposed to the above NLP and ML work, we do not process the text with an automatic algorithm. Instead, the writer's attributes are extracted by human judges through crowd-sourcing, and then serve as features in our prediction algorithm in order to facilitate a high-quality prediction of the author's actions in game-theoretic setups. In Section \ref{sec:approach}, we motivate our crowd-sourcing based approach. We save the automatic extraction of author attributes from the texts, using NLP and ML methods, to future work.

\subsection{Argumentation in Multi-agent Systems}
Our work has several interesting related lines of work carried out in the multi-agent systems community, dealing with argumentation \citep{walton2009argumentation} and negotiation \citep{kraus2001strategic}.
Significant progress on argumentation has been obtained, e.g., by improving prediction of persuasive arguments \citep{azaria2012strategic,azaria2012giving,tan2016winning,rosenfeld2016providing}. Moreover, the long-standing line of research on automated negotiations has developed powerful learning approaches to train automated agents exploiting human choice prediction \citep{PeledGK13,2018Rosenfeldpredi}. Our approach is complementary as its focus is on the task of action prediction from unrelated texts, provided by participants that are unaware of the strategic tasks they will encounter.

Similarly, our work is also complementary to work in multi-agent systems which integrates features of observed social and psychological behavior  in service of agents' creation (e.g., \cite{azaria2012strategic,azaria2012giving,de2014importance,prada2009teaming}) for effective complex interaction with humans. Our work does relate to such theories, but develops the ideas of commonsensical psychological features and focuses on the task of action prediction based on unrelated provided texts.

To illustrate the way our work may complement the above lines of research consider, e.g., \citet{azaria2012giving}. In that work, the authors propose an approach for path selection advice, given by an agent to humans, where the goals of the agent and the humans do not fully coincide. In service of that, an approach integrating machine learning and psychological models for predicting human response is developed. In the web era, many texts generated by people in blogs and reporting their own ``stories'' are available, regardless of a particular coming decision. By having commonsensical features assigned to these texts, as in  our work, one can employ them in service of tasks as the above. Our work provides the first feasibility test of that approach, by having the whole decision relying on texts written by humans while they are not aware of the decision to be made later.

While issues associated with responding to population statistics are quite central to work on human-computer negotiations (e.g., \cite{HaimGGK12}) and human prediction in multi-agent systems applications (see \cite{CranfordLGCVT18}  for multi-agent security applications), the notion of population statistics is used in our setting only as a benchmark for comparison purposes. Our aim is to predict human action in games from the choices made by other humans with ``similar'' attributes, having such statistical information (the majority rule) only as a benchmark.

\section{Task}\label{sec:task}

We now formally describe our choice prediction setup. Let $\mI$ be a set of individuals, and let $\mY$ be a set of action choices (which serve as our labels). The set $\mY$ is composed of all possible choices an individual can make in a given situation. While typical day-to-day decisions correspond to dilemmas, such as whether to buy a product or not or to which academic institute to apply, in this paper $\mY$ serves as the action space of each individual $z\in \mI$ in a normal form, two-player one-shot game that is played between $z$ and another individual, where each one is oblivious to the identity of the other.

We shall assume there is an unknown distribution $\mD_{\mI \times \mY}$ over the product space $\mI \times \mY$. In this formulation, the stochasticity that characterizes decision-making individuals is modeled inside $\mD_{\mI \times \mY}$. More precisely, $\mD_{\mI \times \mY}$ models a mixed action of individuals in the normal form game, but ultimately we assume that each individual makes exactly one decision.

\paragraph{The Games}

In this paper we focus on three (non-cooperative) symmetric two-player games, \textit{Chicken, Box} and \textit{Door}, whose normal-form games appear in Figure \ref{fig:normal form games}. Importantly, we focus on these games, not because individuals encounter them in their everyday activity, but because the dilemmas they capture correspond to real-life situations. In addition, these games reflect a diverse portfolio of dilemmas. $\textit{Chicken}$ is a widely-known game, used as ``a metaphor for a situation where two parties engage in a showdown where they have nothing to gain, and only pride stops them from backing down.''\footnote{\url{https://en.wikipedia.org/wiki/Chicken_(game)}} \textit{Box} is a very simple example of a congestion game \citep{Rosenthal73} with two resources. Notice that congestion games are by all means the most popular type of strategic-form game dealt with in the algorithmic game theory community. \textit{Door}  covers the complementary genre of pure coordination in games, where player payoffs are aligned. These types of games are essential for the study of conventions and focal points \citep{Lewis} in language and beyond. All games are selected to be games with multiple equilibria and no dominant strategies.

\paragraph{Representation}

We are interested in predicting the actions of individuals. To do so, we represent individuals with their personality attributes, where the attribute space is denoted by $\mX$. More concretely, we consider a representation function $\mR:\mI \rightarrow \mX$, such that an individual $z \in \mI$ is represented by $\mR(z) \in \mX$. In our setup $\mR(z)$ corresponds to personality attributes of the participants as extracted from texts they had written prior to participating in the game and without knowing they were going to participate in the game. Details about the texts and the attributes are provided in Section~\ref{sec:data}.
Since individuals are represented according to $\mR$, we shall make the simplifying assumption that each individual $z$ is determined by his/her representation, $\mR(z)$. In that light, we shall address each point $x\in \mX$ as an individual (note that this is with loss of generality), and focus on the induced distribution over representations and choices, $\mD_{\mX \times \mY}$, herein referred to as $\mD$.

\paragraph{Prediction}

A predictive function (or a classifier) $c$ is a function from $\mX$ to $\mY$. A classifier predicts for every individual $x\in \mX$ the action he/she will make, $c(x)$. An empirical evaluation measure $e:(\mY \times \mY)^* \rightarrow \R_{+}$ quantifies the extent to which a sequence of actions $(y)_i$ differs from their predictions $(\hat y)_i$. 

In this paper, we address two action prediction setups: inductive and a transductive, although our focus is on the latter. Both these setups require learning to predict the action of individuals, and the difference between them is in the data available to the algorithm. In the \textit{inductive} setup, a prediction algorithm observes a sample of labeled instances, and has to classify new, unseen instances.  Formally, an inductive learning algorithm is a function from a finite sample to the set of predictive functions, i.e., $\alg: (\mX \times \mY)^* \rightarrow \mY^\mX$, where $\mY^\mX$ is the set of all functions from $\mX$ to $\mY$, i.e., the set of all classifiers. Given a training set $\mS=(x_i,y_i)_{i=1}^m$ of i.i.d. samples from $\mD$, we denote by $\alg^\mS$ the output of $\alg$ on the input $\mS$, which is a predictive function. The score of an inductive algorithm that observes $\mS$ for its training, w.r.t. to a test sample $\mS'=(x_i',y_i')_{i=1}^{m'}$ (drawn i.i.d. from $\mD$ as well) is given by $e\left((y_i',\alg^{\mS}(x_i'))_{i=1}^{\abs{\mS'}}\right)$.

In the \textit{transductive} setup \citep{gammerman1998learning,joachims1999transductive,joachims2003transductive}, a prediction algorithm observes a set of labeled examples along with a set of unlabeled examples, and its task is to predict the labels of the examples in the latter set. Formally, let $\mX' \subset \mX$ be a finite subset of $\mX$. A transductive learning algorithm $\alg$ is a function $\alg: (\mX \times \mY)^* \times (\mX')^* \rightarrow \mY^{\mX'}$. For our purposes, given a sample $\mS'$, let $\mS'\mid_{\mX}$ be the projection of $\mS'$ on $\mX$, i.e., $\mS'\mid_{\mX}=\{x \mid \exists y\in \mY \text{ such that } (x,y)\in \mS'\}$. We denote by $\alg^{\mS,\mS'\mid_{\mX}}$ the output of $\alg$ on the input $\mS,\mS'\mid_{\mX}$, which is a predictive function from $\mS'\mid_{\mX}$ to $\mY$. The score of a transductive learning algorithm is given by $e\left((y_i',\alg^{\mS,\mS'\mid_{\mX}}(x_i'))_{i=1}^{\abs{\mS'}}\right)$.

\begin{figure}[t!]
\centering
\subfigure[]{%
\label{fig:Chicken normal form}%
\begin{tabular}{c|c|c|}
  \multicolumn{1}{c}{} & \multicolumn{1}{c}{Speed}  & \multicolumn{1}{c}{Stop} \\\cline{2-3}
  Speed & $(0,0)$ & $(14,2)$ \\\cline{2-3}
  Stop & $(2,14)$ & $(6,6)$ \\\cline{2-3}
\end{tabular}\quad
}%
\qquad
\subfigure[]{%
\label{fig:Box normal form}%
     \begin{tabular}{c|c|c|}
      \multicolumn{1}{c}{} & \multicolumn{1}{c}{Left box}  & \multicolumn{1}{c}{Right box} \\\cline{2-3}
      Left box & $(8,8)$ & $(16,12)$ \\\cline{2-3}
      Right box & $(12,16)$ & $(6,6)$ \\\cline{2-3}
    \end{tabular}
}%
\qquad
\subfigure[]{%
\label{fig:Door normal form}%
     \begin{tabular}{c|c|c|c|}
      \multicolumn{1}{c}{} & \multicolumn{1}{c}{Door A}  & \multicolumn{1}{c}{Door B} & \multicolumn{1}{c}{Door C} \\\cline{2-4}
      Door A & $(10,10)$ & $(0,0)$   & (0,0)\\\cline{2-4}
      Door B & $(0,0)$   & $(10,10)$ & (0,0) \\\cline{2-4}
      Door C & $(0,0)$   & $(0,0)$   & (8,8) \\\cline{2-4}
    \end{tabular}
}%
\caption{Normal form representation of our games: (a) \textit{Chicken}, (b) \textit{Box}, and (c) \textit{Door}. Each cell in the bi-matrix represents a payoff tuple, where the first entry is the payoff of the row player, and the second entry is the payoff of the column player. The payoffs are given in terms of \textit{points}, which were later converted to US dollars. The games, described in Section \ref{subsec:data collection}, have multiple equilibria and no dominant strategies.}
\label{fig:normal form games}
\end{figure}

\section{Data}
\label{sec:data}

In this section, we describe our data collection process. We describe the process through which we collected personal texts from participants and the games the participants played. Finally, we provide an initial qualitative and quantitative analysis of the collected data. In Section \ref{sec:approach}, we describe the process through which each of the texts was assigned with personal attributes of its author.\footnote{Our data set and the all the information relevant for the data collection crowd-sourcing tasks are publicly available here: \url{https://github.com/omerbp/Predicting-NLPGT}.}

\subsection{Data Collection}\label{subsec:data collection}

\begin{figure}[t]
    \centering
    \includegraphics[scale=0.5]{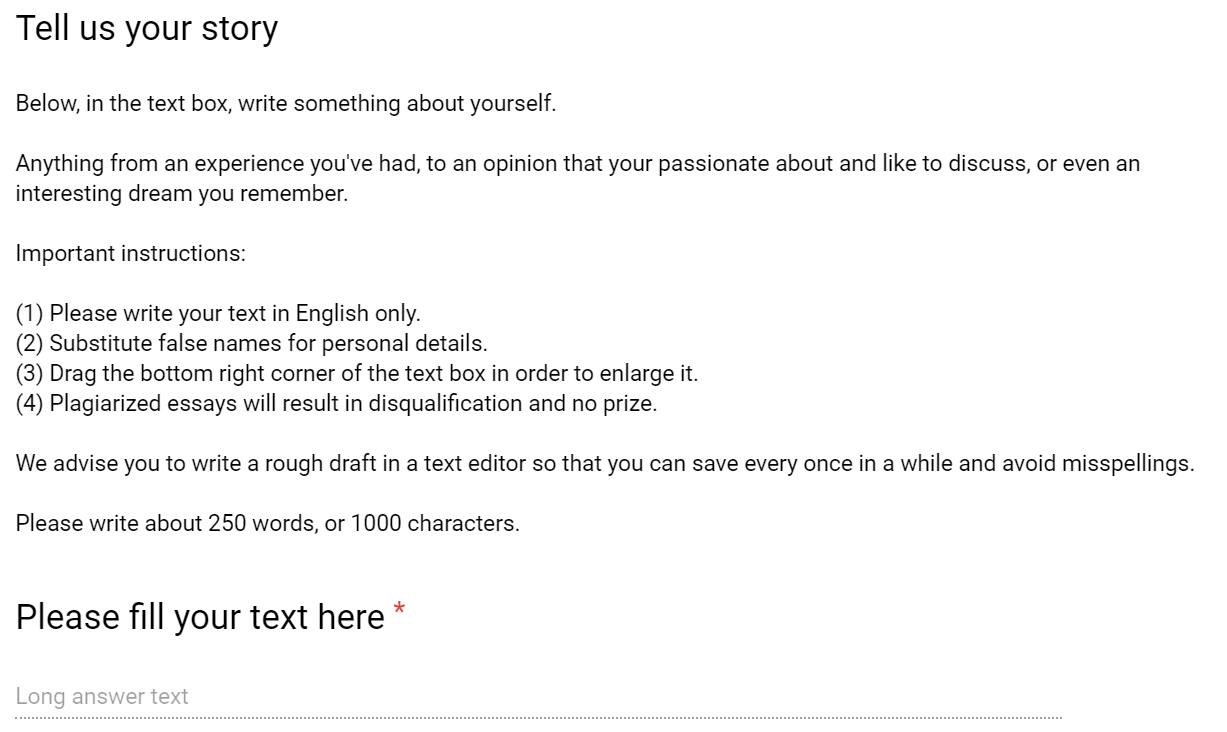}
    \caption{A screen-shot of the writing assignment. The participants were instructed to write a personal text of at least 1000 characters on a topic of their choice.\label{fig:text instructions}}
\end{figure}

The data was collected in two phases: In the first phase, we approached university students with an invitation to participate in an experiment that would grant them a monetary reward. We published an invitation to participate in the experiment on the Facebook page of our university's students, a vibrant page that is used extensively. The reward was declared to consist of compensation for one hour of work (basic payment, see below) plus an additional performance-based bonus. The potential participants (students) were informed that the experiment consists of a short writing task, followed by several short decision-making tasks. Importantly, the students were oblivious to the goals of the experiment. Students who accepted the invitation received an email with a link to the experiment. We shall refer to those as our \textit{participants}.

The experiment was composed of three parts. In the first part, each participant was requested to provide personal information (so that they could be compensated for their participation), along with their gender and age. In the second part, participants were requested to write an English text of at least 1000 characters, on a personal topic of their choice (the instructions are provided in Figure \ref{fig:text instructions}). In the third part, participants were presented with the three non-cooperative games described in Section \ref{sec:task} and Figure \ref{fig:normal form games}, and were told that they are playing with/against another randomly-selected participant. Each game was augmented with a short description of the strategies and payoff functions, and a visualization that depicts the setting. %
The description of the games, after necessary modifications required for inclusion in the paper, is as follows: 

\begin{itemize}
    \item \textit{Chicken}: Each player drives a car, and the two cars are heading towards each other. Each player has the option to Speed or to Stop. In case a player selects Speed, he/she will gain if the other participant selects Stop, but will lose if the other participant selects Speed too. If both participants select Stop, they both gain but to a smaller extent compared to the gain of Speed in the Speed/Stop configuration.
    \item \textit{Box}: Two boxes contain a treasure. Each player selects one box and then obtains either the treasure from that box (if the other player selects the other box) or half of it (in case both players select the same box).
    \item \textit{Door}: There are three doors, A, B and C, and each player should select one of these doors. If the two players select different doors, they gain nothing. If they select the same door, they share the prize associated with that door. Doors A and B have identical prizes, while the prize of door C is slightly lower. 
\end{itemize}

The experiment was online for a time period of one week. We had 280 participants who completed the experiment form. Of these, 9 participants were filtered out, since the text they provided was a copy of a text found on the Web. The total number of participants is hence 271. 
\paragraph{Compensation}

After the data collection phase, we randomly matched the participants,\footnote{Due to the odd number of participants, one participant was matched to herself.} and computed the payoff of each participant. Note that our games do not require simultaneous participation---each participant hence made his/her decision and we computed the payoff according to the random match. Note also, that our prediction tasks are not related to the random match---we predict only the decisions made by the participants, and the random match is computed only for the sake of participant compensation.  

We next converted points to local currency in order to fit our predefined budget and declaration. More precisely, we used a linear mapping from points to the $[10.5,15]$ segment, where $10.5$ was the basic payment. Overall, the average payment was 11.8 US dollars for roughly 40 minutes of work, with a maximum reward of 15 US dollars, a minimum reward of 10.5 US dollars, and a standard deviation of 1 US dollar. The payment reflects the challenging nature of the work: The language of the experiment and the written texts is English, which is not the native language of our participants, bonuses had to be promised to those who excelled (based on their scores in the games), and participants had to come especially to our lab to collect their payment, as the experiment was running online.

\subsection{Quantitative Data Analysis} \label{subsec:Quantitative}

In this section we provide an initial quantitative data analysis. We present the properties of the participants, some of the gathered texts, and statistics of the games. Among the 271 participants, 97 were female and 174 male. We suspect that the bias towards male participants follows from external reasons, e.g., females are underrepresented in engineering schools. The average age was 24, with a standard deviation of three years, a maximum of 32 and a minimum of 18. Further, age highly correlates with gender in the student population in the country of research (Israel), due to external reasons (military service); hence we observe a slight difference in aggregated statistics between genders. The average, median and standard deviation of males were 24.3, 25 and 3.2 years, respectively, compared to 23.7, 24 and 2.3 for females. Our experiment took place in an undergraduate school of a particular university (Technion---Israel Institute of Technology) in a relatively small country, so the country and education are very similar for all participants and hence were not collected. 

\paragraph{The Texts}

The writing instructions presented to the participants guided them to write a personal text on any topic of their choice. The resulting texts addressed topics such as dreams, trips, complaints, among others. To illustrate the quality of these texts, we provide the following representative ones (after making necessary changes to protect the privacy of our participants, and correcting critical typos that would have made the texts difficult the understand). 

\begin{enumerate}
    \item 
	``Hello. My name is Annon. I would like to tell you about a trip I've made to Japan. As most of my good friends, I traveled for a long time abroad - but to a quiet unique destination, Japan. I have always loved Japan - the language, the culture, the history - so I decided to go there. I have travelled for 3 months across Japan, visiting the main island, Honshu, and a few smaller ones- Hokkaido, Shikoku and Kyushu. During my trip I went to the usual attraction, such as Tokyo, Kyoto, and Hiroshima etc. Although those cities are indeed a must, the best experiences for me were in all the small, other places. In Hokkaido I met great people - 3 Iron Man participants, who took me along with them to a tournament that was held in Lake Toya, an amazing women, who I met on a day trip in one of the volcanic mountains, and she took me to unique local places in her car, introduced me to her husband, a hair dresser, and put me in an authentic Kimono while her husband fixed my hair. In Shikoku I danced with the local people and very few foreigners to the Awa Odori festival. In Kyushu I visited a beautiful castle in Kumamoto, danced in an unexpected local festival and travelled with a local friend to a legendary cave surrounded by beautiful river and waterfalls. Overall, I found nothing but pleasant and welcoming people, amazing nature, spectacular festival, delicious food, beautiful gardens and interesting culture and history. I would definitely return soon.''
    \item 
	``During our lifetime, we get to know many people and hear their opinions, yet in my opinion the environment we grow up in influences us the most and shapes our opinions in a certain way that may cause us to be blocked to other people and opinions. A strong opinion that I have matured during my relatively short life, and that I am proud to hold today, is that you have to explore other cultures, get to talk with people that you have never thought of talking to and travel in unfamiliar places. If you create relationships with as many different people as you can, then you are able to better understand their traditions and opinions and become a more tolerant person. In addition, both sides can benefit from hearing what and how the other side thinks and behaves. In my opinion, it is crucial to understand the difference among the different groups in the population you grow up in, and not to be stuck only to a certain line of thought, because stereotypes and prejudice are a weak point that we all sin in it from time to time. Personally, I have explored many cultures and people through traveling around the world, and by creating strong connections with people that not necessarily grew up in my native environment. By that, I think that I have been exposed to many opinions and traditions that have opened my mind to think of other people as much as I think of myself.''
    \item 
	``So a little bit about myself: I'm an architecture student here in the university - and I'm loving it! There isn't much to tell about me. I'm the oldest in my family, I have 3 little sisters and one brother whom I love, and of course my parents. I'm telling you about it because family means a lot to me, I draw power for them and I believe that they do the same. I guess the most important thing I want to pass on is "think pink" or rather "think good" (about people). Lately I've been exposed to so much disrespect and impatience between people. If it because their political opinion or just because they didn't were the right shoes. Often this opinions is hidden beneath a laugh or a sarcastic comment, but we all know that in any laugh there is a bit truth. I believe that words has  power even if you don't mean it those kinds of words and expression are embed in your mind and you don't even notice. Sometimes I hear my friend complain about that girl and that boy and all I can think about is why. If we will put ourselves in the other shoe for a few minute we will realize that most of the people don't want to hurt anyone. I believe that everything has a reason maybe he is just late so you can learn how to be patience.''
    \item 
	``Ever since I was a kid I enjoyed developing my creative skills, whether it's drawing, writing, music or comedy. There's a childhood story about me that my mom likes to tell: when I was three years old we used to live in an old communal apartment in Russia. One day my mother came home from the store and got startled by a large cockroach sitting on the table. As she came closer she realized it was actually a drawing that I made, which apparently seemed realistic enough to fool her. Since then I have continued to try and create more. In middle-school I had a blog, in which I posted random thoughts, stories and Flash animations I made. In high school I used to write short comedic stories and draw comics, and pass them around in class. During that period of time I also learned to play guitar, which soon became another creative outlet. My friends and I used to write a bunch of silly songs, laughing about our school, our surroundings and ourselves. I also wrote a couple of more 'serious' musical compositions, although those still lie buried somewhere between the rest of the files in my computer. When it came to choosing a career path, I was conflicted between engineering and a more artistically inclined path. In the end I chose engineering, knowing that I didn't have to stop creating even though I didn't choose this as a line of work. Today I have a page on Facebook where I post my comics, and I still write the occasional song with my friends.''
    \item 
	``Once I was asked to write a 250 word assay for some sort of an experiment in which I participated. At first I thought that this will be an easy task, as I'm a very opinionated person and I can usually go on and on about practically anything. The problem began when I realized how vague the task was - just write 250 words about anything. How do I choose? Several options came to mind, but I dismissed all of them for different reasons. Eventually I decided to write about that time I was asked to write a 250 word assay. I mean, it's only 250 words, how hard can it be? The first word count showed that I had 119 words. 'Not bad' I thought to myself, as I was almost half way there and the assay even looked more sophisticated and airs poetic then I believed it would. But then, all of the sudden, I hit a writer's block. How do you hit a writer's block writing such a short assay, I thought to myself? But the rhetorical question only helped so much, and I was still stuck with 60 words to go. The good thing about writing an assay with no rules except for 'don't plagiarize' is that you can do practically anything you want. No one will suspect plagiarism if I write something as random as this: the pink elephant believed it was the last of its kind until it realized that it did not in fact exist, and that a different pink elephant was the last of this particular kind.''
\end{enumerate}

\paragraph {Game Statistics}

\begin{figure}[t]
    \centering
    \includegraphics[scale=0.95,valign=t]{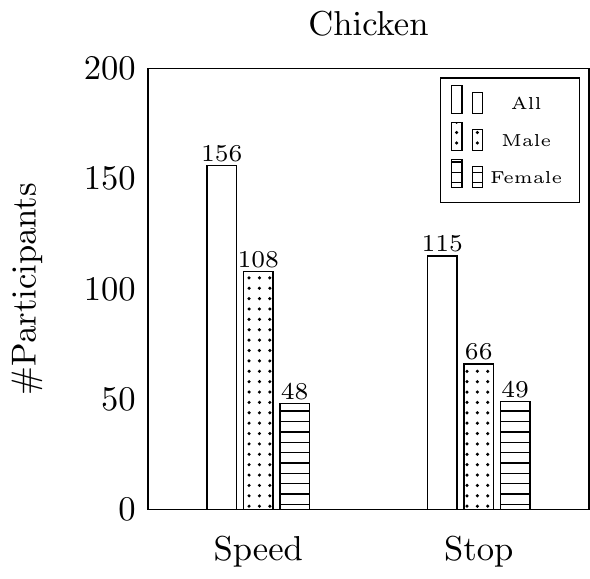}
    \includegraphics[scale=0.95,valign=t]{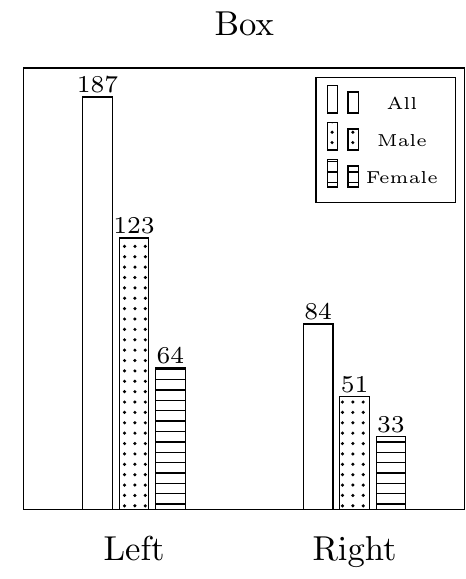}
    \includegraphics[scale=0.95,valign=t]{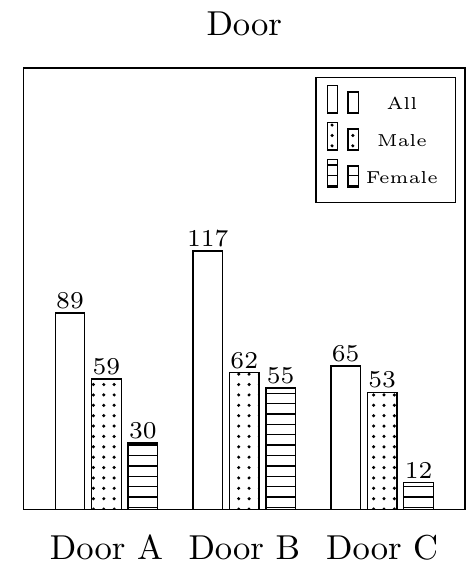}
    \caption{Game play histograms. In \textit{Door}, for instance, 117 participants chose door B, 62 of which are males and 55 are females.\label{fig:games statistics}}
\end{figure}

We now turn to discuss the statistics of the games. This is important not only for the analysis of how our participants behave in the games, but also because our baseline models make use of these simple statistics. The play histograms of all three games are given in Figure \ref{fig:games statistics}. In \textit{Chicken}, the majority (156 participants) selected Speed, while only 115 selected Stop. Interestingly, males are inclined to select Speed, while females are evenly split. In \textit{Box}, 69\% of the participants selected the left box, the one associated with a higher payoff. 

As for \textit{Door}, the payoffs of each door were carefully selected to make door C a focal point, according to the following rationale. If a participant believes her partner aims for the higher reward (obtained by mutual selection of A or B), she still has to guess whether her partner selects A or B, and can be mistaken. Hence, in such a case the participant gets half of $20p$, where $p$ denotes the probability of guessing her partner's strategy correctly between doors A and B. However, if the participant believes her partner aims for the lower reward, she can pick door C and get half of 16. Overall, since the participant and her partner cannot communicate,  if they believe that $p<0.8$ they should pick door C. Nevertheless, the empirical quantities show that the majority of participants selected door B, and some of them explained in hindsight that door B appears in the center, making it, and not door C, the focal point. Interestingly, while males selected the doors almost uniformly, females hardly selected door C. Moreover, more than 56\% of the female participants selected door B in comparison to 36\% of the male participants.

\section{Our Approach}
\label{sec:approach}

We now describe our approach to human choice prediction based on pre-written texts. Our approach is based on two steps: In the first step, we represent each text with the personal attributes of its author. In the second step, we apply a transductive classifier to these examples so that we can predict human choices in games from the choices made by other humans with similar attributes.

\paragraph{Attribute Collection}

In the first step we extract from each text the attributes that we believe are representative of its author. Since we wanted to develop our own set of attributes, we could not rely on existing tools for this extraction task.
Due to the relatively small number of samples, we could not develop a data-driven attribute classifier that would predict our attributes from the texts, and instead we turned to human judgments through a crowd-sourcing platform. This strategy also lets us analyze our approach without having to worry about the noise that is likely to be injected by automatic 
attribute extraction tools. 

\paragraph{Text-based Choice Prediction}

In the second step, we aim to predict the choices made by our participants in a transductive learning setup.
In the transductive setup \citep{gammerman1998learning} one assumes access to a set of examples, but only a subset of these examples is labeled. The goal of a transductive learning algorithm is then to predict the labels of the unlabeled examples. Importantly, all the participating examples are available to the algorithm throughout the process; no new example is introduced at any point, and the prediction task is restricted to the labels of the unlabeled subset of the example set. 

We choose to focus on the transductive setup for both theoretical and practical reasons. Theoretically, in this research we would like to demonstrate the validity of our hypothesis that personal texts written before their authors participate in strategic decision-making setups (games) can help predict the outcome decisions. In real-world setups, a system based on our observations is likely to have access to texts, such as social media posts of a large number of people whose behavior is of interest. It is therefore only reasonable to process these texts in advance so that when the behavior of some authors is being observed, predictions can be made about the others. Practically, our data set was very challenging to collect, as we wanted to make sure we have access to high-quality texts with personal stories, and is hence small (only 271 participants). Transductive learning, which is more effective in small data setups \citep{joachims1999transductive,joachims2003transductive} is hence a natural choice in our case.\footnote{In practice we also performed experiments with several inductive classifiers -- SVM, random forest and xgboost. While development set results were very good, they did not generalize to the test set, which is very common in cases where the data is small.}

Our classifier is based on a deterministic clustering algorithm applied to the attribute-based representation of our participants. Since the clustering does not consider the choices made by the participants in the games, we can derive predictions with respect to all three games from our single, deterministic, clustering. We now elaborate on these two steps.

\subsection{Attribute Collection}\label{subsec:attcollection}

Having collected the data, we turned to extract commonsensical personality attributes from the texts. In service of that, two graduate students read the provided texts and extracted attributes that were considered appropriate as descriptors of the writers. In total, they highlighted 24 attributes, including adventurous, competitive, cynical, introvert, kind, passionate, pessimistic and rational, among others (see the exhaustive list in Table \ref{table:attribute representations}).\footnote{The list of attributes was generated based on all texts. This is in compliance with the transductive setup where all texts are available at training time.} Then, we turned to scoring each text with a number in $\{0,1,\dots 5\}$ for each of these attributes, corresponding to the degree to which the text is affiliated with that attribute. We used Figure Eight (previously known as Crowd Flower)\footnote{\url{https://www.figure-eight.com/}} to extract eight estimates for each text-attribute pair, taking into account the necessary differences between human judgments when it comes to personality questions. 

\begin{figure}[b!]
    \centering
    \includegraphics[scale=0.6]{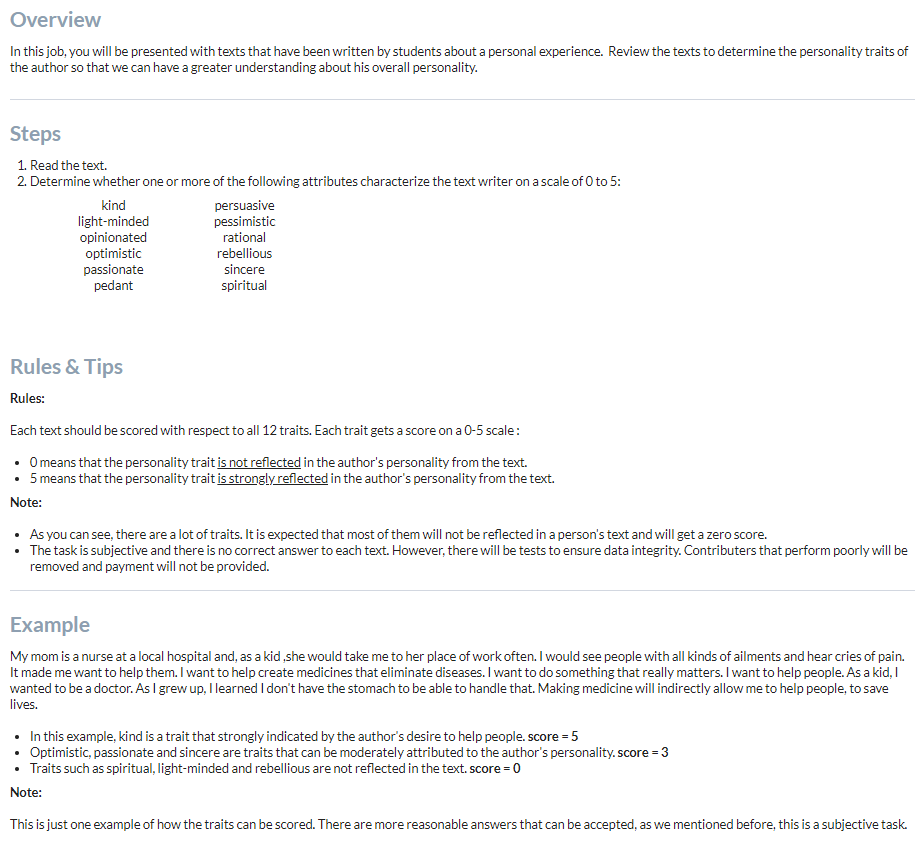}
    \caption{A Screen-shot of the crowd-sourcing attribute collection assignment.\label{fig:attribute instructions}}
\end{figure}

The crowd-sourcing instructions are provided in Figure \ref{fig:attribute instructions}. Note that we do not provide formal definitions for the attributes. Instead, we present the workers with an example and rely on their intuitive understanding of the meaning of these attributes. Hence, controlling the quality of the collected judgments, which is well recognized as an important issue when collecting data through crowd-sourcing (see e.g., \cite{Hill:15} for a recent prominent NLP example), is of particular importance in our setup.

All our crowd-workers were Americans, mostly from New York, California and Virginia. We divided the attributes into two 12-attribute sets, in order to avoid the crowd-workers from experiencing cognitive overload when completing the task. Each crowd-worker observed two texts, and, for each of the texts, had to provide numerical estimations for the attributes in one of the 12-attribute sets. The payment for each task, comprised of reading two texts and providing numerical estimations for 12 attributes, was one US dollar---as recommended by the platform.

To ensure the quality of the collected judgments, each crowd-worker was presented with test questions, selected from a small predefined set that was pre-scored by the authors of this paper. Test questions, which are standard in Human-Computer computation \citep{gormley2010non,Hill:15,bragg2016optimal}, are designed to reduce subjectivity. Such subjectivity is even more challenging in complex semantic interpretation tasks like ours, leading to a low inter-annotator agreement. For example, for word pair similarity \citep{Hill:15}, a semantic interpretation task in NLP, the inter-annotator agreement is lower than 0.7.\footnote{More precisely, the inter-rater agreement is computed as the average of pairwise Spearman correlations (Spearman's rho) between the ratings of all annotators. The overall agreement for three leading datasets, SimLex-999 \citep{Hill:15}, MEN \citep{bruni2012distributional} and WordSim-353 \citep{finkelstein2002placing} is 0.67, 0.68 and 0.61, respectively. Undoubtedly, the task of extracting a single attribute from a medium-size text is more subjective and hence prone to lower agreement rates.\label{footnote:inter-rater}} The test questions could not be distinguished from the other questions, and the system (FigureEight) automatically filtered out crowd-workers that provided unlikely answers to these test questions. Particularly, for each test question we consider an interval of acceptable answers, i.e., 0--2 in case the attribute was not present in the text and 3--5 if the attribute was present in the text, and filtered out workers not complying with at least 70\% success rate on these questions (we emphasize that such workers were instantly filtered by FigureEight, and we set the success rate on 70\% in advance). Overall, we had 770 workers from which 277 (35.9\%) were excluded. The mean scores of the 24 attributes, across all 271 texts, ranging from 0.75 (cynical) to 2.68 (kind) and the std values of these scores range from 1.31 (cynical) to 1.61 (passionate).

Finally, the vector representation of each text was set to the vector whose coordinates correspond to the attributes, and the value in each entry is the average of the scores collected for this text-attribute pair. Table \ref{table:attribute representations} presents the attributes collected for the texts of Section \ref{subsec:Quantitative}. The author of the first text, for example, is most associated with attributes such as adventurous, friendly, kind, lightmineded, optimistic, passionate, and sincere. Naturally, the other authors that did not write about journeys, are not judged to be as adventurous as the author of the first text. Yet, the author of the fourth text, that discusses their personal journey in the career path, is also assigned a relatively high adventurous score.

\begin{table*}[t!]
    \centering
    \begin{tabular}{l c c c c c}%
    Attribute & Text \#1 & Text \#2 & Text \#3 & Text \#4&  Text \#5 
    \begin{filecontents*}{CSV-Files/attribute-table.csv}
attribute,first,second,third,fourth,fifth
adventurous,4.625,1.875,2.125,3.125,1
ambitious,1.875,1.125,1,0.375,1.5
competitive,1.375,1.25,1.375,2.625,2.125
confident,4,3,1.625,2.875,2.625
conformist,1.5,0,1.625,0.375,1.375
creative,1.5,1.75,1,2.875,2.875
cynical,0.25,0.5,2.25,0.75,1.25
friendly,3.75,2.625,2.125,2.625,1.75
humoristic,1.5,1.375,1.125,2.875,1
ideological,2.125,3.125,2,2.5,2.375
insecure,0.375,0,1.75,0.375,0.875
introvert,0.375,0,1,0.375,0
kind,3.625,3.375,4,3.125,0.75
lightminded,3.25,2.75,2.375,1.375,1.25
opinionated,1,0.75,0.25,0.5,3.25
optimistic,3.125,1.5,2.75,3.125,1.125
passionate,3.125,3.875,2.875,4,2.125
pedant,1.125,0,0.375,0,2.25
persuasive,0.875,2.125,1,1.5,0.5
pessimistic,0.625,0,0.5,0.875,1.625
rational,2.375,3.375,2.625,3.125,2.625
rebellious,0.75,1.875,0.375,0.5,2.625
sincere,3.25,3.375,2.75,4,1.875
spiritual,1.125,1.75,1.625,1,0.5
\end{filecontents*}
\csvreader[head to column names]{CSV-Files/attribute-table.csv}{}%
	{\\ \hline\csvcoli&\csvcolii&\csvcoliii&\csvcoliv&\csvcolv&\csvcolvi}
	\\ \hline
    \end{tabular}
    \caption{Attribute values for the texts of Section \ref{subsec:Quantitative}. }
    \label{table:attribute representations}
\end{table*}

\subsection{Attribute-based Transductive Classifier} 
\label{subsec: our model}

We propose a transductive classifier based on a deterministic clustering algorithm. The clustering algorithm is applied to the attribute representation of our participants, and as we are addressing the transductive setup, all the participants are clustered together. Our clustering algorithm does not have access to the choices made by any of our participants, only to their attributes. Hence, we can make predictions with respect to the various games using a single output (clustering) of our algorithm. 

Formally, following the notation of Section \ref{sec:task}, let $\mS$ and $\mS'$ denote an arbitrary partition of the data into a training set and a test set. We assume our algorithm has access to $\mS,\mS'|_{\mX}$, with which it should predict the labels belonging to each element in $\mS'|_{\mX}$. We do this by clustering all the examples in $\mS|_{\mX} \cup \mS'|_{\mX}$, so that we can use the labels of the examples in $\mS$ to make predictions about the labels of the examples in $\mS'$.

In our setup the example set, $\mX$, consists of the entire set of personal texts written by our participants: $\mX = \{x_1, \ldots, x_n \mid  x_j \in \mathbb R^d\}$, where $n = 271$ is the number of our participants and $d=24$ corresponds to the pre-defined writer's personality attributes. Each such $x_j$ corresponds to the vector representation of the $j$-th participant, and each one of the $d=24$ coordinates of $x_j$ corresponds to one attribute and is a real number between 0 and 1 obtained by averaging and re-scaling the scores given to this participant by the crowd-workers with respect to the property in question.
The label set $\mY$ is defined for each game separately, and corresponds to the possible actions of that game.

Our goal is to test the predictive power of text-based personality attributes with respect to a range of human decision-making setups. We hence cluster our example set, $\mX=\mS|_{\mX} \cup \mS'|_{\mX}$, once and then use the resulting clustering to make predictions with respect to all three games. Particularly, we cluster the example set $\mX$ with a bottom-up agglomerative clustering algorithm using Ward's minimum variance criterion for cluster merging \citep{ward1963hierarchical}, which is the default linkage method in the package we employed, Scikit-learn \citep{pedregosa2011scikit}. The output of this process is $n=271$ possible clusterings---from the one at the top of the hierarchical tree, consisting of a single cluster, downward in the tree till the one that consists of $n=271$ clusters. The clustering hierarchy has 271 levels; each level is generated by merging two clusters from its preceding (lower) level in the hierarchy. Given a number of clusters $k$, we obtain the partition of $\mX$ to $k$ clusters from the hierarchical cluster tree and use it for classification as follows: For each $x'\in \mS'|_{\mX}$, i.e., an unlabeled example, we predict its label according to a majority vote between the labeled members of its cluster, i.e., the elements in $\mS|_{\mX}$ which appear in the same cluster as $x'$. In cases of a tie between the number of cluster members with two different labels, or when all the members of a cluster are unlabeled, we make a random prediction about the unlabeled members of the cluster. In what follows, we refer to our algorithm as \textit{TAC}, which stands for Transductive Attribute Clustering.

\section{Experiments}
\label{sec:experiments}

\subsection{Baselines}
\label{sec:baselines}

To measure the quality of TAC, we compare its performance to baselines. Our main baseline is the strong \textit{Majority Vote Classifier (MVC)} that assigns to each example (participant) the majority label in the data. Notice, that this baseline enjoys an advantage over TAC as the majority is computed using all 271 examples. Notice also, that while this is a strong baseline in terms of its overall performance, it fails to predict all classes except from the majority class. Hence, a comparison to this baseline requires a careful selection of evaluation measures. We get back to this point below. 

To put the results of both TAC and MVC in context, we also compute the expected scores of two stochastic classifiers. The \textit{Expected Random Guess (ERG)} score is the expected score of a stochastic classifier that assigns every participant with each of the labels in $\mY$ with probability $\frac{1}{|\mY|}$ (where $|\mY| = 2$ for the \textit{Chicken} and \textit{Box} games, and $|\mY| = 3$ for \textit{Door}). We also compute the expected score of a stronger stochastic classifier---the \textit{Expected Weighted Guess (EWG)} score---which is the expected score of a stochastic classifier that assigns every participant with each of the labels in $Y$ with the probability of that label in the entire example set. Like MVC, this classifier is exposed to the entire labeled set and hence enjoys an advantage over TAC. We consider MVC and EWG to be powerful baselines, because they are exposed to the entire labeled set.

\paragraph{Ablation Analysis}

We further perform ablation analysis to evaluate the importance of each of the components of TAC. The ablation analysis is focused on the importance of the clustering step and the importance of text representation according to the personality attributes of the authors.

To evaluate the importance of the clustering algorithm, we compare our approach to two transductive learning algorithms that do not perform clustering: Transductive SVM\footnote{For the \textit{Door} game that has three classes, we train three pairwise TransSVM classifiers (for each class pair) on the training set, apply each of them to the test set, and set the class of each test set example according to the majority vote of these classifiers, breaking ties randomly.} (\textit{TransSVM}) \citep{joachims1999transductive} with a linear kernel, and a $K$ nearest neighbor (\textit{K-NN}) classifier. Notice that the transductive SVM is trained for each task separately and in this respect it is weaker than TAC that can make predictions with respect to multiple prediction tasks based on its single clustering. Moreover, the transductive SVM takes advantage of the labels of the labeled examples during training, in contrast to TAC that considers only its input examples. The K-NN classifier is identical to TAC except that no clustering is performed and each unlabeled example is labeled according to a majority vote between its $K$ nearest neighbors. 

To evaluate the importance of our character attribute representation, we compare our approach to several models in which we run a clustering algorithm identical to TAC, but with a different feature representation. We employ the following two automatic natural language processing tools:
\begin{itemize}
\item The IBM Personality Insights service.\footnote{ \url{https://personality-insights-demo.ng.bluemix.net/}.} A description of the IBM system, including the psychological research it is based on and its algorithms, is provided at \url{https://cloud.ibm.com/docs/services/personality-insights}. Briefly, the system employs an open-vocabulary approach, building on several recent papers on personality inference \citep{schwartz1992universals,plank2015personality,arnoux201725}. It trains a machine learning algorithm on personality scores from surveys conducted among thousands of Twitter users, and the Twitter feeds of these users, where the texts of Twitter messages are represented with Glove word embeddings \citep{pennington2014glove}. The 13-attribute set of this service consists of the following eight attributes: Joy, Fear, Sadness, Anger, Analytical, Confident, Tentative and Emotional, in addition to the big five personality dimensions \citep{john1991big,costa2008revised}: Extraversion, Agreeableness, Conscientiousness, Neuroticism, and Openness.

\item  Linguistic Inquiry and Word Count (LIWC) \citep{pennebaker2001linguistic,tausczik2010psychological}. LIWC is a popular tool that is heavily used in academic psychological studies. Examples for analyses based on LIWC range from analyzing the language of Al-Qaeda \citep{pennebaker2008computerized} to predicting final course performance of students \citep{robinson2013predicting}. Unlike the IBM's Personality Insights service, which is based on deep neural networks, LIWC is rather simplistic and uses word counts. These word counts are used in order to associate a text with 19 linguistics, psychological and person categories, among which are: Analytic, Clout, Authentic, Power, Money, and Swear. 
\end{itemize}

For each of these tools, we consider two models: One that uses the attributes provided by the tool solely, and another that incorporates both the tool's attributes and our 24 attributes. Particularly, in \textit{TAC-IBM-13} we replace the character attributes we have developed in this work with the ones extracted by the IBM Personality Insights service. We also consider a model named \textit{TAC-IBM-37}, which is identical to TAC but employs both our 24 attributes and the 13 IBM attributes in its text representation. Similarly, \textit{TAC-LIWC-19} is based on the 19 character attributes provided by LIWC, and \textit{TAC-LIWC-43} incorporates both the 19 LIWC attributes and our 24 attributes.

The TAC-IBM-13 and TAC-LIWC-19 models are baselines that put the contribution of our character attribute set in the context of previously-developed and widely-used character attribute sets. The TAC-IBM-37 and TAC-LIWC-43 models help us measure the extent to which our character attribute set is complementary to these previously proposed character sets, and to quantify the added impact of our set.

In the last baseline we consider, the feature representation of each example is the tf-idf bag-of-words representation of the text. That is, this model, which we denote as \textit{TransTextCluster}, uses a word-based text representation rather than the personality attributes extracted for the author.

\subsection{Classifier Evaluation}\label{subsub:eval}

Several of the aforementioned models are classifiers that learn from a train set of labeled examples and draw predictions on a test set of unlabeled examples: TAC (and its variants: TAC-IBM-13, TAC-IBM-37, TAC-LIWC-19, and  TAC-LIWC-3),  TransTextCluster, TransSVM and K-NN. In order to evaluate these algorithms, we perform for each game the following procedure 5000 times: We randomly sample train and test sets, $\mS$ and $\mS'$, such that the training set is comprised of 90\% of the data. Then, for each of the unlabeled examples, namely $x'\in \mS'|_{\mX}$, we predict a label. For TAC and TransTextCluster, we do that  according to the process described in Section \ref{subsec: our model}, while prediction with TransSVM and K-NN is described in Section \ref{sec:baselines}.  The number we report for each (algorithm, game) pair is the average of the evaluation measure values across the 5000 repetitions. 

For TAC we focus our evaluation on the range of 2--30 clusters. For K-NN we consider $K \in \{1, \ldots, 5\}$. For the clustering with tf-idf representation, we consider 2--30 clusters, as for TAC, and compute tf-idf for the 1904 vocabulary words after removing stop words and punctuation marks. We next discuss the evaluation measures. 

\subsection{Evaluation Measures}

Our data demonstrates a class imbalance phenomenon, where in each game one of the actions is observed substantially more frequently than the other(s) (Figure \ref{fig:games statistics}). We hence consider three complementary evaluation measures that can help us tell the full story about our model, the baselines and the expected values of the random classifiers.

In class imbalance situations like ours, it is important to evaluate the ability of an algorithm to successfully detect the various classes in the data, be they large or small. For this aim, we turn to a standard solution and employ the Macro Average F1-score. For this measure, we compute the F1-score of each class and report the average of the resulting values. More specifically, let $y_j$ be the label of the $j$'th example in the test set $\mS'$, and let $\hat y_j$ be the predicted label of that example. For each class $i\in \mY$, we compute
\begin{equation*}
Precision_{i} = \frac{\sum_{j=1}^{|\mS'|} \ind_{\hat y_j = y_j=i}}{\sum_{j=1}^{|\mS'|} \ind_{\hat y_j=i}}, \quad
Recall_{i} = \frac{\sum_{j=1}^{|\mS'|} \ind_{\hat y_j = y_j=i}}{\sum_{j=1}^{|\mS'|} \ind_{y_j=i}}, \quad
F1_{i} = \frac{2 \cdot Recall_{i} \cdot Precision_{i}}{Recall_{i} + Precision_{i}}.
\end{equation*}
In other words, $Precision_{i}$ is the fraction of instances belonging to class $i$ from those classified as class $i$, while $Recall_{i}$ is the fraction of instances from class $i$ classified as class $i$. The $F1_i$ score is the harmonic average of the precision and recall of the $i$-th class. The Macro Average F1-score is:
\begin{equation*}
MAV-F1 = \frac{1}{|\mY|}\sum_{i=1}^{|\mY|} F1_{i},
\end{equation*}
where $|\mY|$, the number of classes is equal to 2 for \textit{Chicken} and \textit{Box}, and to 3 for \textit{Door}. Informally, $MAV-F1$ is the average of the class-based $F1$ scores, with equally weighted classes. To take the class size into account more directly, we also consider the Macro Weighted Average F1 score:
\begin{equation} \nonumber
MWAV-F1 = \sum_{i=1}^{\abs{\mY}} \sum_{j=1}^{|\mS'|} \frac{\ind_{y_j =i}}{|\mS'|} \cdot  F1_{i}.
\end{equation}

Finally, we consider the Accuracy measure, defined as the fraction of test set examples that are correctly labeled by the algorithm:
\begin{equation} \nonumber
Accuracy = \sum_{j=1}^{|\mS'|} \frac{\ind_{y_j =\hat y_j}}{|\mS'|}.
\end{equation}

\section{Results}
\label{sec:results}
\begin{sidewaysfigure}
    \centering
    \includegraphics[scale=.75]{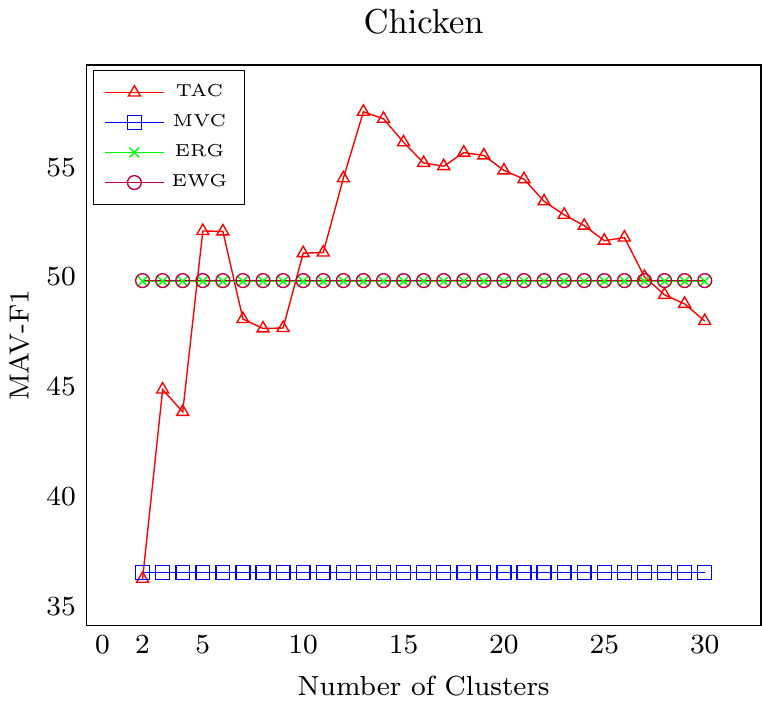}
    \includegraphics[scale=.75]{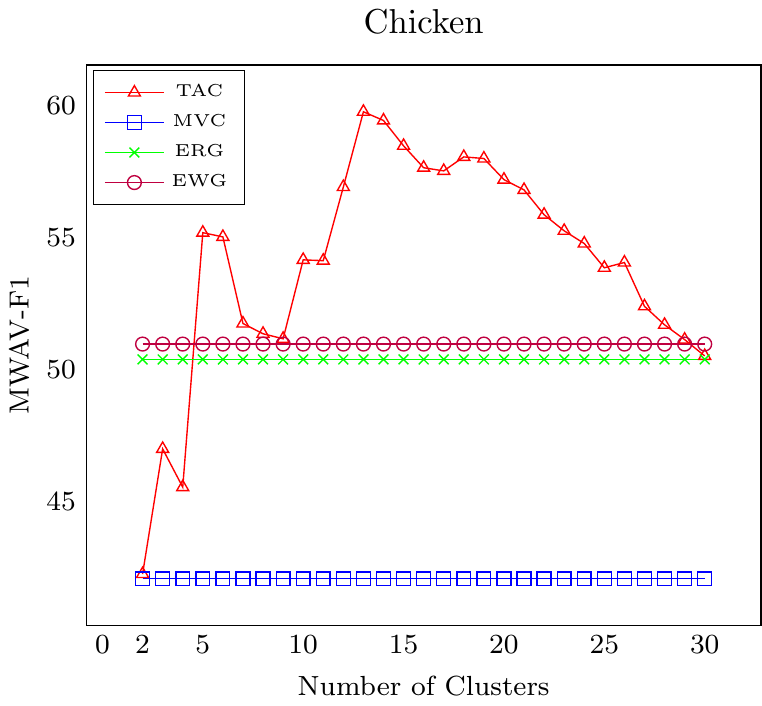}
    \includegraphics[scale=.75]{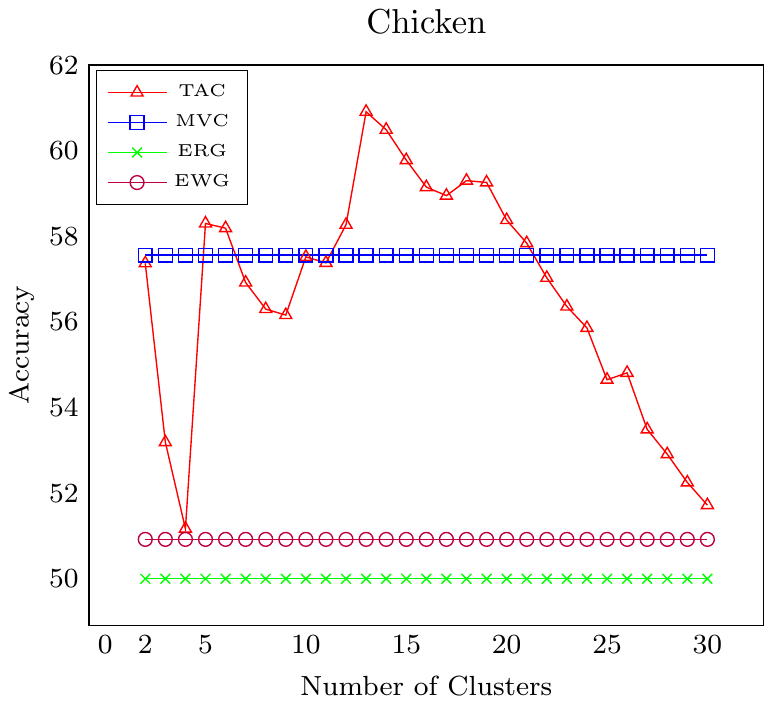}
    \includegraphics[scale=.75]{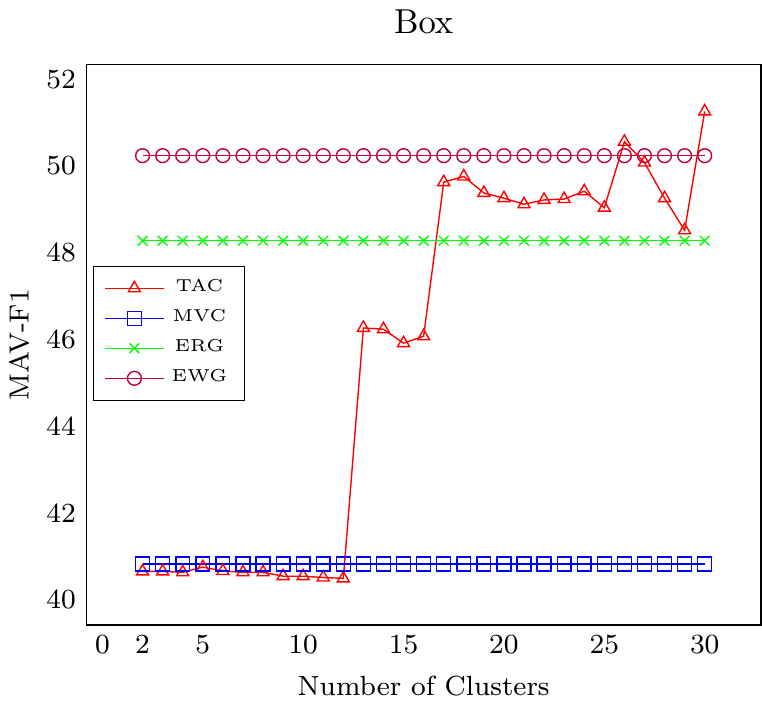}
    \includegraphics[scale=.75]{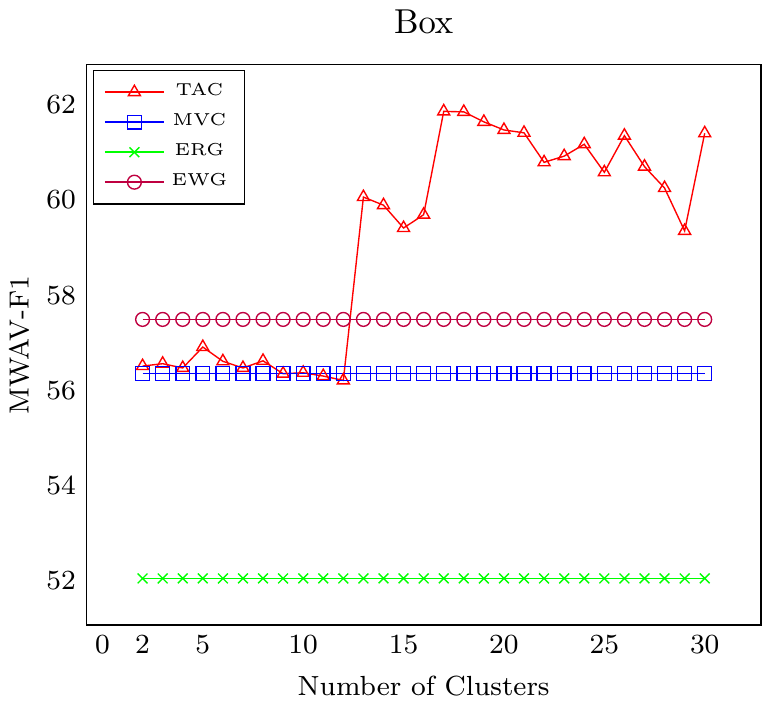}
    \includegraphics[scale=.75]{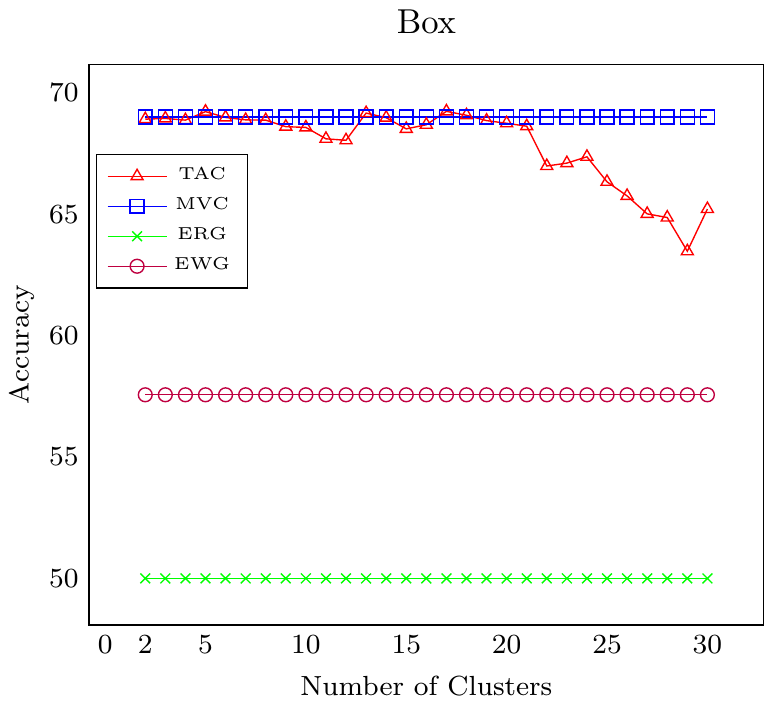}
    \includegraphics[scale=.75]{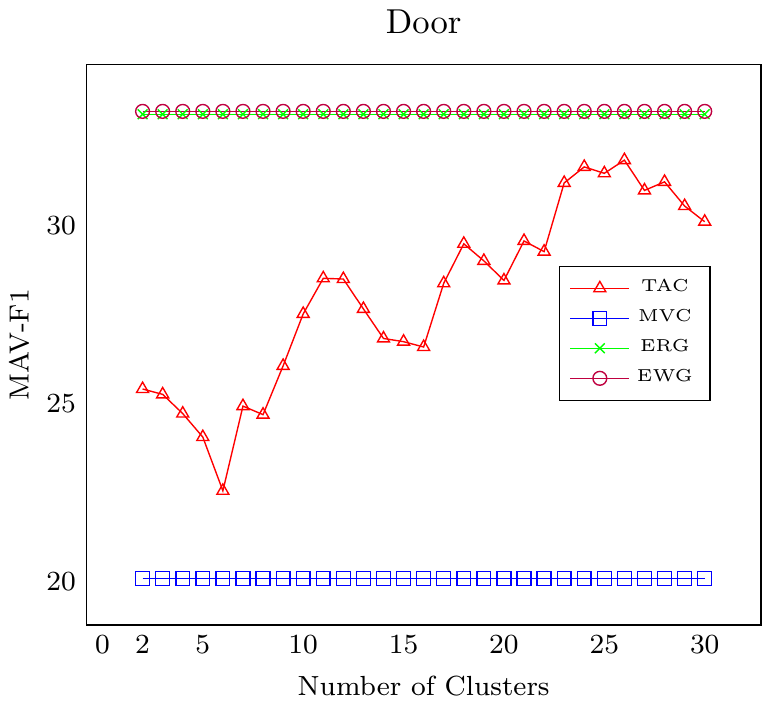}
    \includegraphics[scale=.75]{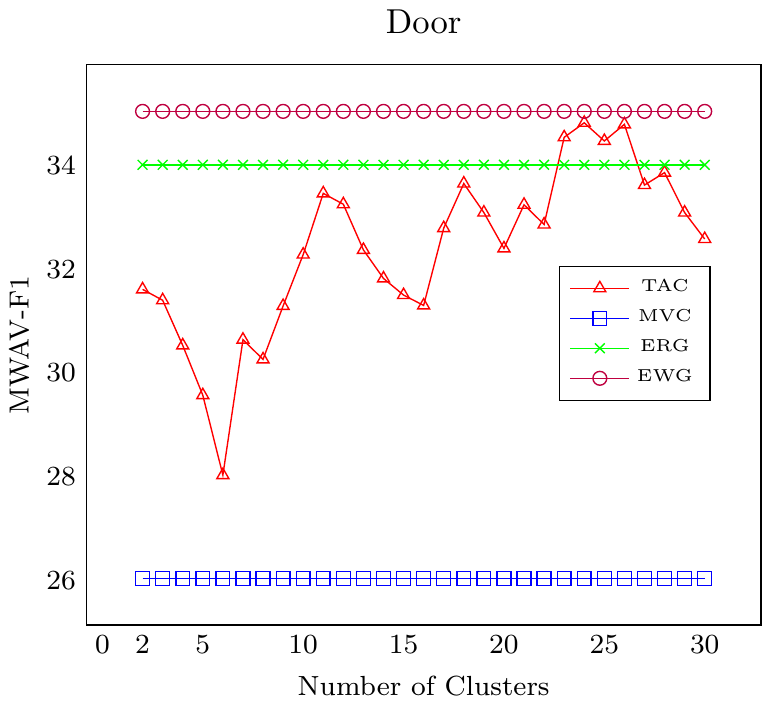}
    \includegraphics[scale=.75]{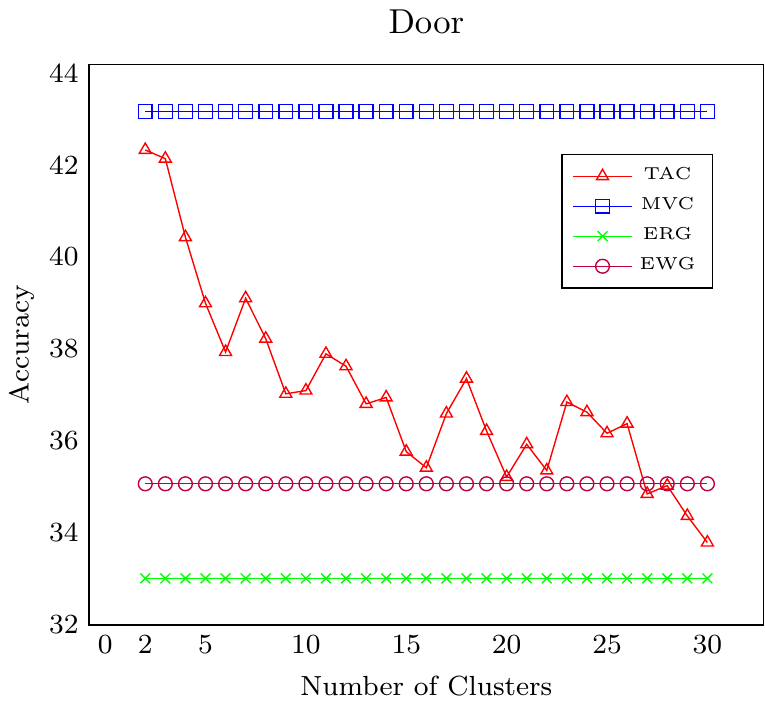}
    \caption{Results for TAC, the MVC baseline and the ERG and EWC expected scores.}
    \label{fig:results}
\end{sidewaysfigure}

Figure \ref{fig:results} presents the Macro Average F1, Macro Weighted Average F1 and Accuracy measures for our clustering algorithm (TAC), the majority vote classifier (MVC) and the two random classifiers (ERG and EWC). We note that the reported values for ERG and EWC are expectations rather than scores of one specific prediction and they hence only serve for putting the results of the two actual classifiers, TAC and the MVC baseline, in context.\footnote{As noted in Subsection \ref{subsub:eval}, we performed 5000 repetitions for each estimate. Chernoff-like concentration inequalities guarantee that using that many repetitions ensures negligible confidence intervals and error bars. \label{footnote:sig}
}
\paragraph{Classifier Comparison: TAC vs. MVC}

We first compare the two classifiers: TAC and MVC. When considering the Macro Average F1 measure, that assigns an equal weight to the classes regardless of their size, TAC outperforms MVC. For \textit{Chicken} and \textit{Door} this result holds regardless of the number of clusters TAC produces, while for \textit{Box} it holds when TAC generates 13 clusters or more. Moreover, the same pattern is observed when considering the Macro Weighted Average F1, that weighs the per-class F1 scores according to the class size. For Accuracy, a measure that is strongly affected by success on the majority class, MVC outperforms TAC for \textit{Door}, but TAC performs better for \textit{Chicken}, starting with 10 clusters, and both algorithms perform very similarly for \textit{Box}.

This pattern clearly demonstrates the superiority of TAC. While it is hard to beat an MVC algorithm on Accuracy, TAC clearly detects both the minority and the majority classes well enough so that it strongly outperforms MVC on the Macro Average F1 measures. Moreover, even on Accuracy, MVC does not enjoy a clear advantage over TAC. Remembering that TAC performs clustering once, in a task-ignorant manner, and uses the resulting clustering to perform predictions for multiple tasks, the advantage of our algorithm is even more impressive.

Table \ref{table:f1} provides another demonstration of the favorable performance of TAC. The table presents the per class F1 score for the various algorithms, where for TAC it considers the performance for the configuration that yields the best MAV-F1 results (13 clusters for \textit{Chicken}, 30 for \textit{Box} and 26 for  \textit{Door}). While MVC, which labels all examples with the majority class, should naturally perform better on the majority class, its F1 score on that class is better than those of TAC by only  4.99, 4.88 and 13.56 for \textit{Chicken}, \textit{Box} and \textit{Door}, respectively. On the minority classes TAC enjoys nice F1 score gaps from MVC whose F1 score for these classes is 0 by definition.

\paragraph{Comparison with Expected Values of Random Classifiers}

To put our results in context, we also compare the performance of our algorithm to the expected performance of a random classifier that assigns a participant with a class sampled from a uniform distribution over classes (ERG) and with the expected performance of a random classifier that samples the assigned class from the class distribution of the data set (EWG). Table \ref{table:f1} demonstrates that the F1 score of the majority class is substantially higher for TAC in all three games, while the opposite trend is observed when considering the minority classes in \textit{Box} and \textit{Door}. For \textit{Chicken}  TAC performs best even on the minority class. This pattern is reasonable, as it is much easier to get \textit{expected} high scores on the minority class rather than to commit to a \textit{single} high-quality prediction on them.

Figure \ref{fig:results} presents the global measures that result from this fine-grained pattern. For Macro Average F1 and Macro Weighted Average F1, TAC is competitive with the expected scores of the random classifiers. For \textit{Chicken}, this is mostly true for 10--25 clusters, while for \textit{Box} and \textit{Door}, TAC generally improves as more clusters are produced. For Accuracy, TAC is clearly better, regardless of the number of clusters. Due to the class imbalance (Figure \ref{fig:games statistics}), our task is best evaluated with a combination of measures. TAC clearly provides a better balance between the various measures compared to the expected scores of the random classifiers.

\begin{table*}
        \small
    \begin{tabular}{l| c c| c c| c c c }%
    &\multicolumn{2}{c|}{\bfseries Chicken} & \multicolumn{2}{c|}{\bfseries Box} &  \multicolumn{3}{c}{\bfseries Door}\\
\hline Algorithm&\textit{F1-Speed}&F1-Stop&\textit{F1-Left}&F1-Right&F1-Door A&\textit{F1-Door B}&F1-Door C\\
\hline TAC&68.08&\textbf{46.97}&76.78&25.74&24.50&46.75&24.29\\
\hline MVC&\textbf{73.07}&0&\textbf{81.66}&0&0&\textbf{60.31}&0\\
\hline ERG&53.60&46&58.20&\textbf{38.36}&\textbf{33.52}&37.5&\textbf{28.39}\\
\hline EWG&57.23&42.42&69.33&31.14&32.18&43.22&24.24\\
\hline
    \end{tabular}
     \caption{Per-class F1-score for each of the classes in our three games. The  majority class of each game is highlighted in italics, and the best value for each measure is highlighted in bold. TAC results refer to the number of clusters that yields the best MAV-F1 results: 13 for \textit{Chicken}, 30 for \textit{Box} and 26 for  \textit{Door}. 
 \label{table:f1}}
\end{table*}

\begin{table*}[t!]
    \centering
    \footnotesize
    \setlength\tabcolsep{2.5pt} 
    \begin{tabular}{l | c c c |c c c| c c c}%
    &\multicolumn{3}{c|}{\bfseries Chicken} & \multicolumn{3}{c|}{\bfseries Box} &  \multicolumn{3}{c}{\bfseries Door}
\\
\hline Algorithm&Acc.&MAV-F1&MWAV-F1&Acc.&MAV-F1&MWAV-F1&Acc.&MAV-F1&MWAV-F1 \\
\hline TAC&\textbf{61.2}&\textbf{57.82}&\textbf{60.05}&65.26&51.4&61.43&36.5&31.95&34.95\\
\hline K-NN&46.95&45.05&46.91&57.85&51.44&58.52&33.42&30.1&33.28\\
\hline TransSVM&50.24&48.76&50.64&53.40&45.37&53.84& 35.00& 32.91& 35.45\\
\hline TAC-IBM-13 & 53.53 &	48.65 &	51.41 &  65.68 &	41.22 &	56.05 &  41.54 &	36.85 &	39.39 \\  
\hline TAC-IBM-37 & 57.66 &	51.44 &	54.27 & 66.21 &	\textbf{51.49} &	\textbf{61.93} & 42.51 &	37.95 &	40.59	 \\ 
\hline TAC-LIWC-19 & 51.92 & 42.32 & 46.23 &  \textbf{68.14} &40.74 & 56.43 &  \textbf{46.67} &	\textbf{38.50} & \textbf{43.04}	 \\
\hline TAC-LIWC-43 & 55.39 & 51.27 &53.83 &  64.9 &	42.88 & 56.75 &  40.02 &	35.81 &	38.62 \\
\hline TransTextCluster &51.92&37.45&41.95&67.76&42.19&57.12&41.43&27.37&32.91\\ \hline
    \end{tabular}
    
\vspace{1cm}

    \begin{tabular}{l | c c c |c c c| c c c}%
    &\multicolumn{3}{c|}{\bfseries Chicken} & \multicolumn{3}{c|}{\bfseries Box} &  \multicolumn{3}{c}{\bfseries Door}

\\
\hline Algorithm&Acc.&MAV-F1&MWAV-F1&Acc.&MAV-F1&MWAV-F1&Acc.&MAV-F1&MWAV-F1 \\
\hline TAC&  \textbf{58.18} & 	\textbf{52.05} & 	\textbf{54.99} & 	\textbf{68.96} & 	46.21 & 	\textbf{59.81} & 	36.68 & 	28.53 &	32.94 \\
\hline K-NN& 52.97 & 	43.72 & 	47.64 & 	61.76 & 	\textbf{50.21} & 	59.62 & 	36.80 & 	30.03 & 	34.27 \\
\hline TransSVM& 50.24& 	48.76 &	50.64 & 	53.40 & 	45.37 & 	53.84 &	35.00 & 	32.91 & 	35.45\\
\hline TAC-IBM-13 & 51.07 & 	41.99 & 	45.61 & 	68.78 & 	40.62 & 	56.51 & 	38.14 & 	33.10 & 	36.00 \\ 
\hline TAC-IBM-37 & 53.42 & 	43.90 &	47.69 & 	63.70 & 	44.74 & 	57.13 & 	39.73 & 	33.55 & 	36.24	 \\ 
\hline TAC-LIWC-19 & 47.33 &	40.41 &	43.7 &  66.9 &	40.46 & 56.03 &  \textbf{44.27} &	\textbf{36.31} &	\textbf{41.11} \\
\hline TAC-LIWC-43 & 56.71 & 48.25 &  51.83 &  65.22 &	41.31 & 55.67 &  37.00 &	28.92 &	33.37 \\
\hline TransTextCluster & 56.68 & 	36.13 & 	41.99 & 	68.86 & 	40.64 &	56.51 & 	38.92 & 	23.47 & 	28.62  \\ \hline
    \end{tabular}
    
    \caption{Ablation analysis. K-NN and TransSVM use the same personality attribute features as TAC, while TAC-IBM-13, TAC-IBM-37, TAC-LIWC-19, TAC-LIWC-43 and TransTextCluster use the same clustering algorithm as TAC, but with the 13 IBM or 19 LIWC personality attributes (TAC-IBM-13 and TAC-LIWC-19, respectively), both the IBM attributes and ours (TAC-IBM-37), both the LIWC attributes and ours (TAC-LIWC-43), or with TF-IDF textual features (TransTextCluster). For TAC, K-NN, TAC-IBM-13, TAC-IBM-37, TAC-LIWC-19, TAC-LIWC-43 and TransTextCluster the upper table  reports results with their best MAV-F1 hyper-parameter value, while the lower table reports results for the hyper-parameter value that provides the median MAV-F1 values (TransSVM does not depend on a hyper-parameter, so its results are identical in both tables). Numbers of clusters/neighbors for each algorithm in each of the tables are reported in the text.}
    \label{table:ablation} 
\end{table*}

\paragraph{Ablation Analysis}

Our final comparison aims to evaluate the importance of the two main components of TAC: the clustering algorithm and the personality attribute representation. As discussed in Section \ref{sec:baselines}, in order to understand the importance of the clustering, we compare to TransSVM, a transductive parametric classifier, and to K-NN, a transductive parameter-free classifier. In order to understand the impact of the attribute representation, we compare to TAC-IBM-13, TAC-IBM-37, TAC-LIWC-19, TAC-LIWC-43 and TransTextCluster that are identical to TAC in their transductive clustering-based classifier, but differ in the feature set employed for text representation: TransTextCluster represents the texts with tf-idf bag-of-word features, TAC utilizes our 24 personality attributes set, TAC-IBM-13 utilizes the IBM 13 personality attributes set, TAC-IBM-37 utilizes the attributes of both our set and IBM's, TAC-LIWC-19 utilizes the 19 LIWC personality attributes set, and TAC-LIWC-43 utilizes the attributes of both our set and the LIWC set.

Table \ref{table:ablation} presents the results for the hyper-parameter configuration that yields the best (upper table)\footnote{Hyper-parameter values that give the best results (upper table) are: K-NN: (1, 1, 1) neighbors, TAC: (13, 30, 26) clusters, TAC-IBM-13: (30, 25, 8) clusters, TAC-IBM-37: (4, 23, 19) clusters, TAC-LIWC-19: (17, 11, 14) clusters, TAC-LIWC-43: (28, 20, 30) clusters, and TransTextCluster: (28, 25, 28) clusters, for (\textit{Chicken}, \textit{Box}, \textit{Door}), respectively.} or median (lower table)\footnote{Hyper-parameter values that give the median results (lower table) are: KNN: (2, 5, 5) neighbors, TAC: (6, 13, 17) clusters, TAC-IBM-13: (11, 9, 18) clusters, TAC-IBM-37: (12, 16, 9) clusters, TAC-LIWC-19: (30, 15, 24) clusters, TAC-LIWC-43: (9, 16, 13) clusters, and TransTextCluster: (11, 12, 11) clusters, for (\textit{Chicken}, \textit{Box}, \textit{Door}), respectively.} test set\footnote{We note that the K-NN, TAC, TAC-IBM-13, TAC-IBM-37, TAC-LIWC-19, TAC-LIWC-43 and TransTextCluster results in Tables \ref{table:f1} and \ref{table:ablation} are presented for the number of clusters (TAC, TAC-IBM-13, TAC-IBM-37, TAC-LIWC-19, TAC-LIWC-43 and TransTextCluster) or neighbors (K-NN) that yield the best or median test-set results, among the various values considered for these parameters (2--30 clusters, 1--5 neighbors). Recall that due to the small size of our data set we did not use a development set and hence, in order to present two representative numbers for each algorithm per evaluation measure, we chose to focus on the hyper-parameter values that yield the best and median results of each algorithm. A full comparison between TAC, the random classifiers and MVC is provided in Figure \ref{fig:results}. For the ablation analysis, we focus on two representative numbers in order to avoid  cognitive load.} MAV-F1 result for each model. For \textit{Chicken}, TAC is the best performing model, when considering both the median and the best results. For \textit{Box}, when considering the best result, TAC-IBM-37, which utilizes both our attributes and the IBM attributes, yields the best MAV-F1 and MWAV-F1. Note that TAC (with our attributes) does much better than TAC-IBM-13 (with the IBM attributes) on these measures. When considering the median results, TAC achieves the best accuracy and MWAV-F1, and is second only to K-NN (which uses our 24 personality attributes) on MAV-F1. TAC models with LIWC features (TAC-LIWC-19 and TAC-LIWC-43) are overall not competitive for \textit{Chicken} and  \textit{Box}, with the exception of the accuracy measure for \textit{Box}.

For \textit{Door}, both for the best and the median, TAC-IBM-13 performs better than TAC, but TAC-IBM-37 yields an additional performance gain and is the second-best performing model. Yet, for this game TAC-LIWC-19 is the best model. 

As to a feature set combination, our attribute set has an additive effect with the IBM attribute set for all three games (compare the TAC-IBM-37 and TACL-IBM-13 lines in both tables for all games). In contrast, the LIWC set does not add well with ours, which is particularly noticeable in \textit{Chicken} and \textit{Door} when one of the feature sets is very effective (ours in \textit{Chicken}, LIWC in \textit{Door}) and the combined model, TAC-LIWC-43 is strongly outperformed by the TAC model with the single effective attribute set (TAC for \textit{Chicken} and \textit{Box}, TAC-LIWC-19 for \textit{Door}).

Overall, these results justify our clustering and representation choices. Particularly, our clustering algorithm is utilized by all winning configurations (with the single exception of K-NN yielding the strongest median result for \textit{Box}). Our attribute set is utilized by the winning configurations of two of the three games (\textit{Chicken} and \textit{Box}), and by the second best configuration of the third game (\textit{Door}). Given the many years and thousands of studies where the IBM and LIWC attributes have been in use, our novel commonsensical approach to personality attribute generation and annotation performs surprisingly well.

\section{Discussion}
\label{sec:discussion}

Our work initiates a study of action prediction in a single one-shot game. In contrast to previous studies, we do not have access to agents' behavior or to population statistics in other games. Instead, we have access to texts provided by the agents (the participants in our experiments), and we exploit these texts and other agents' behavior in the targeted game in service of action prediction.
We note that while our work does not deal with an ensemble of games, we also tested whether one can predict the play in one game based on behavior in the other games, given our data about the three games,  but this led to no significant result.

On a technical level, although our study is based on a relatively large data set, compared to controlled experiments in previous studies, it is still too small to predict the actions of individuals using heavy inductive methods such as deep neural networks. An important next step would be to generate larger data sets in service of our task, and to further validate our approach. 

Our work focuses on the transductive learning setup where at training time the learning algorithm has access to the texts written by all the participants, including those for which a prediction should be made at test time. As discussed in Section \ref{sec:approach}, this approach is realistic in many real-world setups since a system based on our observations is likely to have access to texts, such as social media posts of a large number of people whose behavior is to be predicted.  A larger data set would also allow us to better understand the extent to which the success of our approach relies on the transductive setup, and whether modern inductive algorithms can compensate for the lack of textual data from the agents whose behavior is to be predicted, at the training time of the algorithm.

In this research, texts are represented with commonsensical personality attributes that are defined by two graduate students (without any special background in psychology) and assigned to the text by crowd-workers. Another important line of future work has to do with a better understanding of these commonsensical attributes that have been demonstrated to provide useful signals about human behavior. We are also interested in a deeper understanding of the relations between our attribute set and previously proposed sets, also in contexts other than our action prediction task. Ideally, such research may also result in a further improvement of our attribute set as well as previous ones, and in better
understanding which attributes convey predictive power with respect to different action prediction tasks. Finally, once we collect a larger data set we can also try to learn the attributes automatically from text using text classifiers, in order to implement our approach in a fully-automatic system.

On a more conceptual level, an interesting follow-up research would be to try and predict an opponent's action in a game when they play after seeing the agent's text. This will make our study closer to the cheap talk literature mentioned in the introduction. We comment though that when agents are aware that their texts are part of a game, as in the cheap talk literature,  then they may choose their texts strategically. In our current study we took care that texts will not be written strategically, as they were isolated from the game play. This made the prediction task highly challenging, leading to what we see as surprisingly positive results.
\subsection{Potential Limitations of this Study}
As is often the case with applied research, our study raises thoughts and comments on the data collection process, the validity of the results, and on replicability.  Despite our efforts to promise a clean experimental plan and execution, some aspects may still be elusive and raise controversies. We devote this section to discuss such potential limitations.

First, our data collection process injected biases w.r.t. gender and age. Since we have decided to collect the data in our university, such biases are inevitable due to the unique environment of our participants (university students). Unfortunately, females are still under-represented in technical universities; the age group of students is a characteristic of our society. 

Second, extracting the list of personality attributes (see Subsection \ref{subsec:attcollection}) was done in isolation from the other parts of the experiment. The graduate students who authored this list passed the authors only the \textit{list} of attributes, and not the representation of each text with these attributes (i.e., the features of each text). Next, text annotation with this list of attributes was done through crowd-sourcing. We have generated a handful of new texts and annotated them ourselves with attributes from the list, so that they can serve as instructions for this crowd-sourcing task. We further created score intervals for test questions to ensure that crowd-workers are making a reasonable effort. Notice that as these are new texts that were not part of our strategic decision experiment and served only for crowd-worker instructions, they are not augmented with any information about strategic behavior. 

One may claim that since the authors of this paper are aware of the attribute use in the strategic prediction, by creating the instruction page and setting the score intervals in the test questions, we might have biased the Figure Eight crowd-workers. However, we could not induce anything from the list of attributes before the texts were annotated with the attributes by the Figure Eight crowd-workers, particularly whether some attributes reveal strategic behavior. Hence, we could not push crowd-workers towards specific personality attribute annotations that reveal strategic behavior. 

To strengthen the last point, if there were such a straightforward connection between attributes and strategic behavior, the performance measures for our algorithm (e.g., Accuracy) would have been much higher. To that end, recall that non-sparse feature vectors of 24 dimensions (see Table 1) and a highly non-linear classification method (TAC, Transductive Attribute Clustering) yielded encouraging results, but these results are still far from perfect. We cannot rule out that our instructions and generated examples had \textit{any} effect on the crowd-workers, but we are confident that the potential effect was not overwhelming. 

Third, we have employed a filtering process during attribute collection. As discussed in Subsection \ref{subsec:attcollection} and then in Footnote \ref{footnote:inter-rater}, these are common techniques in tasks with low inter-annotator agreement. Undoubtedly, the task of extracting personality attributes from a medium-size text is more subjective than typical sentiment annotation for texts such as movie reviews, and is hence prone to lower agreement rates. Since the connection between text and strategic behavior is not at all clear at first glance, we believe that reducing the noise in the attribute collection processes is crucial. Furthermore, the filtering was part of the logic of the attribute-based text annotation process: Once a worker erred on more than 30\% of the questions, he/she was asked to stop filling out the form, and their annotations were left partial.

Additional relevant issues are replicability and significance. It is clear to us that a human experiment, like the one we ran with the crowd-workers, is not replicable in the sense that we cannot guarantee that if someone runs the same protocol he/she will get the same results (personality attribute annotation of texts). However, this study is replicable in the sense that one can run the protocol in the same way we did, as this paper provides the exact details of all the steps we took. Moreover, since we employed data quality assurance measures, we can hope that the data we collected reflect something more general about people's judgment that will be replicated in new experiments. Our experiment is hence replicable when considering the accepted standards of the psychological literature: Researchers can re-run our experiment, including the personality attribute generation process, and compare their results to ours. To emphasize this, we made available online the entire data: The texts, the actions, and the attribute representation. This way, not only could our results be replicated, but they could also potentially be improved.

As for significance, the fact that Figure \ref{fig:results} and Tables \ref{table:f1} and \ref{table:ablation} include neither error bounds nor significance levels follow from our boosting practice, as mentioned in Footnote \ref{footnote:sig}. The analysis is repeated 5000 times, differing in the random split of subjects into test and train sets. We can thus guarantee that our results are significant. 

To facilitate replicability and significance even further, in future work we intend to annotate a large set of personal texts (e.g., social media posts) with our attributes. We will then see if we can learn these attributes with an automatic classifier, and whether the results we get with these automatically-learned attributes are similar to what we report in the current paper. If the answers are positive, then our results would become replicable in the CS sense of the word as well. We did not do that in the present work, as we considered this process to be beyond our scope.

Finally, we review our approach of collecting commonsensical features for the prediction task versus using existing NLP tools. Recall that, as described in Table \ref{table:ablation}, features generated by two existing psychology-based NLP tools, LIWC and the IBM Personality Insights service, were often outperformed by our approach. While analyzing the features obtained from these tools, we realized that they produced fairly accurate estimates of personality attributes (at least according to our subjective judgment), but those were not strongly indicative of chosen game strategies. We hence turned to designing our own set of attributes, as described throughout the paper. However, by no means do we argue that no NLP tool could succeed on our prediction task, and it is not unlikely that feature engineering from multiple existing automatic tools could provide even better results than those we report for the IBM and LIWC feature sets.

On the bright side, we find the fact that two students with no extensive psychological background extracted features that are indicative of our choice prediction task, to be a major strength rather than a shortcoming. Generalizing, this suggests that commonsensical features could be useful for human behavior prediction. The psychology literature presents a wide gamut of well-established personality characterizations, yet our approach suggests that relying on expert knowledge from the psychological literature is not always required.

\clearpage
\section*{Acknowledgments}
The authors wish to thank the editor and the anonymous reviewers for their helpful feedback and questions, which led to substantial improvement of this paper. We thank Hily Abramovich and Nadav Oved for extracting commonsensical personality attributes from the texts (as explained in Subsection \ref{subsec:attcollection}), and Idan Sugarman for assisting in the data collection process (as explained in Subsection \ref{subsec:data collection}). In addition, we would like to thank the members of the IE@Technion NLP group for their valuable feedback and advice. The work of O. Ben-Porat is partially funded by a PhD fellowship from JPMorgan Chase \& Co. The work of M. Tennenholtz is funded by the European Research Council (ERC) under the European Union's Horizon 2020 research and innovation programme (grant agreement n$\degree$  740435).

\bibliographystyle{theapa}

\end{document}